\documentclass{bmvc2k}

\title{Bayesian Geodesic Regression on Riemannian Manifolds}

\addauthor{Youshan Zhang}{yoz217@lehigh.edu}{1}

\addinstitution{
Computer Science and Engineering\\
Lehigh University\\
Bethlehem, PA, USA
}

\runninghead{Zhang Y.}{Bayesian Geodesic Regression on Riemannian Manifolds}

\usepackage{microtype}
\usepackage{graphicx}
\usepackage{subfigure}
\usepackage{booktabs} 
\usepackage{mathrsfs}  
\usepackage{amssymb,amsmath}
\usepackage{hyperref}
\usepackage{wrapfig}
\usepackage{cutwin}
\usepackage{algorithm}  
\usepackage{stmaryrd}
\usepackage{algorithmic}
\usepackage{amsfonts,amssymb}
\usepackage{verbatim}
\usepackage{sidecap}
\usepackage{color}
\usepackage{url}
\DeclareMathOperator{\Exp}{Exp}
\DeclareMathOperator{\Log}{Log}

\usepackage{multirow}

\begin{document}

\maketitle

\begin{abstract}
 Geodesic regression has been proposed for fitting the geodesic curve. However, it cannot automatically choose the dimensionality of data. In this paper, we develop a Bayesian geodesic regression model on Riemannian manifolds (BGRM) model. To avoid the overfitting problem, we add a regularization term to control the effectiveness of the model. To automatically select the dimensionality, we develop a prior for the geodesic regression model, which can automatically select the number of relevant dimensions by driving unnecessary tangent vectors to zero. To show the validation of our model, we first apply it in the 3D synthetic sphere and 2D pentagon data. We then demonstrate the effectiveness of our model in reducing the dimensionality and analyzing shape variations of human corpus callosum and mandible data.

\end{abstract}

\section{Introduction}
Regression analysis is a predictive modeling technique, which studies the relationship between dependent variables (objectives) and independent variables (predictors). This technique is usually used to predict and analyze the causal relationship between variables. The benefits of regression analysis are numerous, such as:  (1) it shows the significant correlation between the independent variable and dependent variable; (2) it shows the influence of multiple independent variables on a dependent variable.
Regression analysis also allows us to compare the interactions between variables that measure different scales.

However, linear models are not applicable if the response variable takes the values on the  Riemannian manifold. Manifold learning has been applied in many fields, including domain adaption, transformation, tensor and shape measurement \cite{gopalan2011domain,gong2012geodesic,zhang2019transductive,rathi2007segmenting,pizer1999segmentation,zhang2019mixture}. However, it has difficulty in analyzing the shape variations, which are essentially high-dimensional and nonlinear. Therefore, it is necessary to develop a general regression model and reduce the dimensionality on manifolds.

Several studies have explored the regression issues on the manifolds, which includes unrolling method, regression analysis on the group of diffeomorphisms, nonparametric regression, second order splines, a semiparametric model with multiple covariates, and geodesic regression ~\cite{jupp1987fitting,miller2004computational,davis2010population,trouve2010second,shi2009intrinsic,fletcher2011geodesic,zhang2019k}. However, these methods does not provide a Bayesian framework for the generalization of geodesic regression on manifolds. It is thus necessary to develop such a model to automatically choose the model complexity. 

The purpose of this paper is to develop a generalized Bayesian geodesic regression on Riemannian manifolds, termed BGRM model. Our model can estimate the relationship between an independent scalar variable and a dependent manifold-valued random variable. Our work is inspired by the Bayesian principal component analysis (BPCA) model, which is introduced in Euclidean space by Bishop~\cite{bishop1999bayesian}.

We develop a maximum likelihood posterior model for Bayesian geodesic regression on manifolds (BGRM). By introducing a prior to the geodesic regression model, we can automatically select the number of relevant dimensions by driving unnecessary tangent vectors to zero. The main advantage of our Bayesian geodesic regression approach is that the model is fully generative. The unnecessary dimensionality of the subspace will be automatically killed, and the principal models of variation can reconstruct shape deformation of individuals. To show the validation of our model, we first apply it in the 3D synthetic sphere and 2D pentagon data.  We then use the human corpus callosum and mandible data to show the predicted shapes using our model. Our results indicate that the BGRM model provides a better description of the data than PCA~\cite{jolliffe2011principal}, principal geodesic analysis (PGA)~\cite{fletcher2004principal} and probabilistic principal geodesic analysis (PPGA)~\cite{zhang2013probabilistic} estimations. Our model also shows reasonable shape variations with the increasing of age in a much lower-dimensional subspace.

\section{Bayesian linear regression on Euclidean space (BLR)}

\begin{figure}[ht]
\centering
\subfigure[Linear regression]{\label{fig:BLR}
\includegraphics[width=0.45\columnwidth]{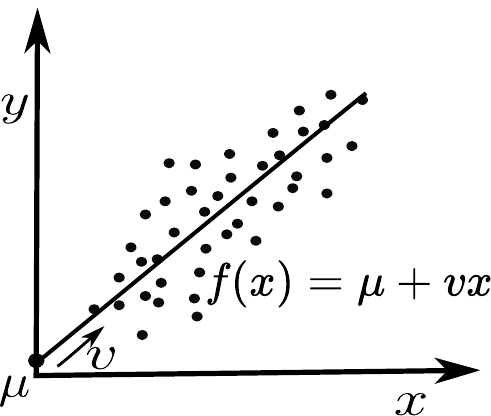}
}
\centering
\subfigure[Geodesic regression]{\label{fig:BGRM_model}
\includegraphics[width=0.48\columnwidth]{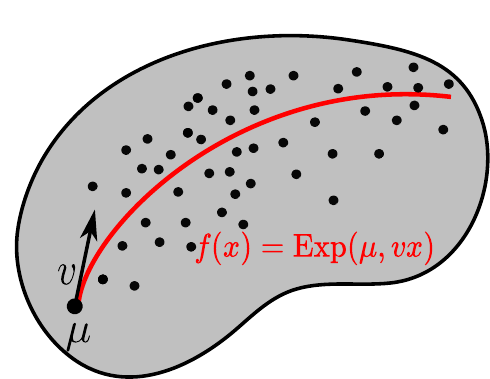}
}
\caption{(a): Schematic of the linear regression, $\mu$ is one point (the intercept), $v$ is the tangent vector (the slope). The black line is the linear regression line $f(x) = \mu + vx$.  (b): Schematic of the  geodesic regression on Riemannian manifolds, $\mu$ is a base point on the manifold, $v$ is a point of tangent space. The red line is the geodesic regression line $f(x) = \Exp(\mu, vx)$. }
\label{fig:models}
\end{figure}

Before formulating Bayesian geodesic regression on Riemannian manifolds, we first review Bayesian linear regression on Euclidean space. Fig.~\ref{fig:BLR} shows the scheme of linear regression model. Given the target variable (dependent variable) $y \in \mathbb{R}^{n} $, and independent variable $x \in \mathbb{R}$. The linear regression model is given by:
\begin{equation}\label{lg}
    y= \mu + v x  +\epsilon,
\end{equation}

where $v \in \mathbb{R}^n$ is the unobservable \textit{slope} parameter, $\mu \in \mathbb{R}^{n}$ is the unobservable \textit{intercept} parameter, and $\epsilon$ is zero mean Gaussian unobservable random variable with the variance $\tau^{-1}$. Thus, we can rewrite Eq.~\eqref{lg} as:


\begin{equation}\label{nlg}
    p(y|x,\mu,v,\tau)=\mathcal{N}(y|\mu + v x, \tau^{-1}), 
\end{equation}

where $p (\cdot)$ is the probability, and $\mathcal{N}$ is the normal distribution. Consider a data set of input $X=\{x_n\}_{n=1}^{N}$ ($x_{n} \in \mathbb{R}$) with corresponding target value $Y=\{y_{n}\}_{n=1}^{N}$ ($y_{n} \in \mathbb{R}^{n}$ can be treated as column vector with the size of $d \times 1$), and these data points are independently sampled from the normal distribution. Then, the data likelihood is:

\begin{equation}\label{log}
    p(Y|X,\mu,v,\tau)=\prod_{n=1}^{N}\mathcal{N}(y_{n}|\mu + v x_{n}, \tau^{-1}). 
\end{equation}

To automatically select the principal component, Bayesian geodesic regression model includes a Gaussian prior over each column of $v$, which is known as an automatic relevance determination (ARD) prior. The $v$ is constrained to a zero-mean isotropic Gaussian distribution with the parameter $\alpha$:
\begin{equation}
    p(v|\alpha)=\mathcal{N}(v|0,\alpha^{-1}I).
\end{equation}

The logarithm of posterior distribution is given by (see \cite{bishop1999bayesian} for details):

\begin{equation} \label{lnre}
    \text{ln}\  p(v|y)=-\frac{\tau}{2}\sum_{n=1}^{N}\{y_{n}-(\mu+vx_{n})\}^{2}-\frac{\alpha}{2}v^{T}v + \text{const}..
\end{equation}

The value of $\alpha$ is iteratively estimated  as $\alpha=\frac{N}{v^{T}v}$, if $\alpha_i$ (one column of $\alpha$) is large, the corresponding $v_i$ will be small and thus enforces sparsity by driving the $v_{i}$ to zero. The sparsity of $v$ has the same effect as removing irrelevant dimensions in the principal subspace.

\section{Bayesian geodesic regression on manifolds (BGRM)}

\subsection{Background: Riemannian Geometry }

In this section, we recap three essential concepts (Geodesic, Exponential, and Logarithmic Map) on the Riemannian Geometry (more details are  provided by~\cite{zhang2013probabilistic,zhang2019mixture,zhang2019transductive}).

\textbf{Geodesic.} Let $(M, g)$ be a Riemannian manifold, where $g$ is a Riemannian metric on the manifold $M$.
Consider a curve  $C(t): [0,1] \rightarrow M$  and let $C'(t) = dC/dt$ be its velocity.  We call $C$ a geodesic if $C'(t)$ is parallel along $C$, that is: $C''=\frac{d C'}{dt}= \nabla_{C'} C'=0$, which means the acceleration vector (directional derivative) $C''$ is normal to $T_{C(t)} M$ (the tangent space of $M$ at $C(t)$).  Note that geodesics are straight lines in Euclidean space ($\mathbb{R}^{n}$).

\textbf{Exponential Map.} For any point $p \in M$ and its tangent vector $v$, let $\mathcal{D}(p)$ be the open subset of $T_{p} M$ defined by: $\mathcal{D} (p) = \{v \in T_{p} M| C(1)\}$, where $C$ is the unique geodesic with initial conditions $C(0) = p$ and $C'(0) = v$. The \textit{exponential map} is the map $\Exp_{p} : \mathcal{D} (p) \rightarrow M $ defined by: $\text{Exp}_{p}(vt_{=1}) = C(1)$, which means the exponential map returns the points at $C(1)$ when $t=1$. $\text{Exp}_{p}(vt)$ can also be denoted as: $\text{Exp}(p, vt)$. In Euclidean space, the exponential map is the addition operation $\text{Exp}_{p}(vt) = p + vt$.

\textbf{Logarithmic Map.} Given two points $p$ and $p'$ $\in M$, 
the \textit{logarithmic map} takes the point pair $(p, p')$ and maps them into the tangent space $T_{p} M$, and it is an inverse of the exponential map: $\Log(p, p') \rightarrow T_{p} M$. $\Log(p, p')$ can also be denoted as: $\Log_{p} p'$. Because Log is an inverse of the exponential map, we can also write: $p'= \Exp(p, \Log (p, p'))$. The \textit{Riemannian distance} is defined as $d(p, p') = \| \text{Log}_{p}(p')\|$. In  Euclidean space, the logarithmic map is the subtraction operation: $\text{Log}_{p}(p') = p' - p$.

\subsection{Geodesic regression}
Geodesic regression has been proposed by Fletcher~\cite{fletcher2011geodesic}, the geodesic regression model is defined as:
\begin{equation}
    Y=\Exp(\Exp(\mu,vX),\epsilon),
\end{equation}

where $\mu$ is a base point on the manifold, $v$ is a point of tangent space $T_{\mu} M$, $X$ is the independent variable, $Y$ is the observed data and $\epsilon$ is a random variable taking values in the tangent space with the precision $\tau$. Since the exponential map is the addition operation in Euclidean space, the geodesic regression model coincides with Eq.~\eqref{lg} when $M = R^{n}$. Fig.~\ref{fig:BGRM_model} shows the scheme of geodesic regression model.

Eq.~\eqref{eq:Nr} is the Riemannian normal distribution  $\mathcal{N}_M(\mu, \tau^{-1})$,  with its precision parameter $\tau$.  This general distribution  can be applied to any Riemannian manifold (see~\cite{zhang2013probabilistic} for details). 
\begin{align}\label{eq:Nr}
p(y | \mu, \tau) &= \frac{1}{C(\mu, \tau)} \exp\left( - \frac{\tau}{2} d(y, \mu)^2 \right), \quad \text{where} \\ \nonumber
C(\mu, \tau) &= \int_M \exp\left( -\frac{\tau}{2} d(y, \mu)^2 \right) dy.
\end{align}

Given a data set of input $X=\{x_n\}_{n=1}^{N}$ ($x_{n} \in \mathbb{R}$) with corresponding target value $Y=\{y_{n}\}_{n=1}^{N}$ ($y_{n} \in \mathbb{R}^{n}$ can be treated as column vector with the size of $d \times 1$) on general manifolds. Each target value $y_n$ is independent of the Riemannian normal distribution. Therefore, the data likelihood on Riemannian manifolds is defined as:

\begin{equation}\label{anlg}
    p(Y|X,\mu,v,\tau)=\prod_{n=1}^{N}\mathcal{N}_M(y_{n}|\Exp(\mu, v x_{n}), \tau^{-1}). 
\end{equation}

Taking the logarithm of the Eq.~\eqref{anlg}, we have

\begin{align}\label{M_loglg}
    \text{ln} \ p(Y|X,\mu,v,\tau)&=-N \ \text{ln} \ C -     \frac{\tau}{2}\sum_{n=1}^{N}\{\Log(y_{n},\Exp(\mu, v x_{n}))\}^{2}.
\end{align}

\subsection{Regularized geodesic regression}

The overfitting problem can be caused by a complex model that is trained on a small number of datasets.  To avoid the overfitting problem, we add a regularization term in Eq.~\eqref{M_loglg}, thus the total energy function (E) is given by:

\begin{equation} \label{M_error}
\begin{aligned}
E &= \frac{1}{2}\sum_{n=1}^{N} \{\Log(y_{n},\Exp(\mu, v x_{n}))\}^{2}+\frac{\gamma}{2}v^{T}v,
\end{aligned}
\end{equation}

Choosing the optimal dimensionality is significant, we need to find the appropriate number of basis functions to determine a suitable value of the regularization coefficient $\gamma$. Therefore, it is necessary to develop a method which can automatically choose the dimensionality of data.

\subsection{Bayesian geodesic regression}

To automatically select the principal geodesic
from data, we also include a Gaussian prior over each column of $v$.  Supposing that $v$ is constrained to a zero-mean isotropic Gaussian distribution, which has a single parameter $\alpha$, so that:
\begin{equation}
    p(v|\alpha)=\mathcal{N}(v|0,\alpha^{-1}I).
\end{equation}

Since the posterior distribution $p(v|y)$ is proportion to $p(y|x,v,\tau) \times p(v|\alpha)$, the logarithm of posterior distribution is equivalent to take the logarithm of $p(y|x,v,\tau) \times p(v|\alpha)$. Then the log of the data likelihood can be computed as:

\begin{equation}
\begin{aligned}\label{lnmre}
    &\mathbb{E}=\text{ln} p(v|y)=- N \ \text{ln} \ C -\frac{\tau}{2}\sum_{n=1}^{N}\{\Log(y_{n},\Exp(\mu, v x_{n}))\}^{2}-\frac{\alpha}{2}v^{T}v + \text{const.}.
\end{aligned} 
\end{equation}


Maximization of this posterior distribution with respect to $v$ is equivalent to the minimization of the sum-of-squares error function with the addition of a quadratic regularization term, corresponding to Eq.~\eqref{M_error} with $\gamma=\alpha/ \tau $.

\begin{figure}[h]
\centering
\includegraphics[scale=0.4]{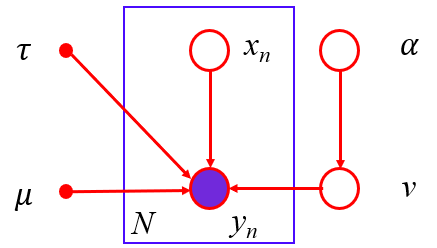}
\vspace{+0.3cm}
\caption{Probabilistic graphical representation of BGRM model. In the blue box, it has $N$ data points, for each independent observation $y_n$, it is together with the corresponding independent variable $x_n$, and it is associated with $\tau, \mu$ and $v$. Here, $v$ is controlled by the hyperparameter $\alpha$. }
\label{fig:graphic_model}
\end{figure}

Similar to the BLR model, the value of $\alpha$ is iteratively estimated  as $\alpha=\frac{N}{v^{T}v}$, and then enforces sparsity by driving the corresponding component $v_{i}$ to zero. More specifically, if $\alpha_{i}$ is large, $v_{i}$ will be effectively removed. This arises naturally because the larger $\alpha_{i}$ is, the lower probability of $v_{i}$  will be. Therefore, we can automatically select the dimensionality of $v$, and it has the same effect as removing irrelevant dimensions in the principal subspace. Fig.~\ref{fig:graphic_model} is the graphical representation of BGRM model.

\subsubsection{Gradient terms}
We use gradient descent to maximize the posterior distribution in Eq.~\eqref{lnmre}, and update the parameters $\mu, v$ and $\tau$. However, the computation of the gradient term requires us compute the derivative of $\Exp(vx_n, \mu)$, which can be separated into a derivation with respect to the initial point $\mu$, and  respect to the initial velocity $v$ (please refer to \cite{fletcher2011geodesic} for the details of derivations).

Now we are able to take the gradient of Eq.~\eqref{lnmre} with respect to different parameters: $\mu, v$ and $\tau$ and get the following gradient terms:

\textbf{Gradient for $\mu$:} the gradient of $\mu$ is

\begin{equation}\label{eq:gmu}
\begin{aligned}
\nabla_{\mu}(\mathbb{E})=-\sum_{n=1}^{N} \tau d_{\mu}\text{Exp}(\mu, v x_{n})^\dag \text{Log}(y_n,\text{Exp}(\mu, v x_{n})),
\end{aligned}
\end{equation}
where $\dagger$ represents adjoint operation, for any  $u$ and $w$
$$\langle d_{\mu} \Exp( \mu, v x_{n})u, w \rangle = \langle u, d_{\mu} \Exp( \mu, v x_{n})^{\dagger} w\rangle.$$

\textbf{Gradient for $v$:} the gradient of $v$ can be computed as

\begin{equation}\label{eq:glambda}
\begin{aligned}
\nabla_{v}(\mathbb{E})&=- \sum_{n=1}^{N}  \tau x_{n} d_{v}\text{Exp}(\mu, v x_{n})^\dag \text{Log}(y_n,\text{Exp}(\mu, v x_{n}))-\alpha v.
\end{aligned}
\end{equation}

{\bf Gradient for $\tau$ :} the gradient of $\tau$ is computed as

\begin{equation}\label{eq:gtau}
\begin{aligned}
\nabla_{\tau}(\mathbb{E}_{\mathcal{L}})&=\sum_{n=1}^{N}  \frac{1}{C(\tau)}A_{n-1}\int_{0}^{R}\frac{r^2}{2}\text{Exp}(-\frac{\tau}{2}r^2)\prod_{k=2}^{n}k_{k}^{-1/2}\times\\&f_{k}(\sqrt{k_k}r)dr -\frac{1}{2}\{\Log(y_{n},\Exp(\mu, v x_{n}))\}^{2}dr,
\end{aligned}
\end{equation}
where $A_{n-1}$ is the surface area of $n-1$ hypershpere. $r$ is radius, $k_{k}$ is the sectional curvature. $R=\text{min}_{v}{R(v)} $, and $R(v)$ is the maximum distance of $\text{Exp}(\mu, rv)$, $v$ is a point of unit sphere $S^{n-1} \subset T_{\mu} M$. However, this formula only validates for simple connected symmetric spaces, other spaces should be changed according to the different definitions of the probability density function (PDF) in Eq.~\eqref{eq:Nr} (please see \cite{zhang2013probabilistic} for details).

\section{Evaluation}

\subsection{Data}
{\bf Sphere.} To validate our model on the spherical manifold, we simulate a random sample of $293$ data points (3D) on a unit sphere with known parameters (ground truth) in Table \ref{tab:my_label}. The points are randomly sampled given the $\mu$, $v$, and the precision $\tau$ (please refer to~\cite{fletcher2016probabilistic} for generating points on a sphere).  The ground truth $\mu$ is generated from random uniform points on the sphere, and $v$ is generated from a random Gaussian matrix.
\begin{table}[b]
    \caption{Parameters comparison between ground truth and the estimation of BGRM, PGA and PCA models ($[\cdot]^T$ is the transpose of a matrix).  }
    \vspace{+0.3cm}
    \label{tab:my_label}
    \centering
    \newcommand{\tabincell}[2]{\begin{tabular}{@{}#1@{}}#2\end{tabular}}
   \setlength{\tabcolsep}{+2mm}{
    \begin{tabular}{c c c c }
    \hline
            &  $\mu$   & $v$ & $\tau$\\
         \hline
        Ground truth  & $[0.7704, 0.4155, 0.4836]^T$& 
 $\left[
 \begin{matrix}
   0.0755  & -0.2771 &  -0.2784 \\
   -0.0002  &  0.0007 &   0.0100 \\
  \end{matrix}
  \right]^T  $ & 100 \\
        \hline
        BGRM    & $[0.8728, 0.2916, 0.3913]^T$ & 
         $\left[
 \begin{matrix}
   0.0739  & -0.2577  & -0.2239 \\
  \end{matrix}
  \right]^T  $ & 98.3399 \\
        \hline
        PPGA \cite{zhang2013probabilistic}  & $[0.8674, 0.2735, 0.4157]^T$ & 
   $\left[
 \begin{matrix}
   0.0773  &  -0.2849   & -0.1873 \\
    -0.0006  & 0.0012  &  0.0098 \\
  \end{matrix}
  \right]^T  $       & 
        103.817 \\
        \hline
        PGA~\cite{fletcher2004principal}  & $[0.8848, 0.2850, 0.3687]^T$ & 
        
   $\left[
 \begin{matrix}
   0.0752  &  -0.1070   & -0.0978 \\
    -0.0076  & 0.0748  &  0.0761 \\
  \end{matrix}
  \right]^T  $       & 
        N/A \\
        \hline
        PCA~\cite{jolliffe2011principal} & $[0.8680, 0.2797, 0.3618]^T$ & 
        
           $\left[
 \begin{matrix}
   0.0717  &  -0.1065  &  -0.0979 \\
    -0.0083  & 0.0746  &  0.0751 \\
  \end{matrix}
  \right] ^T $   &  N/A \\
        \hline
    \end{tabular}}
\end{table}

\begin{figure}[t]
\centering
\subfigure[]{\label{fig:sphere}
\includegraphics[width=0.3\columnwidth]{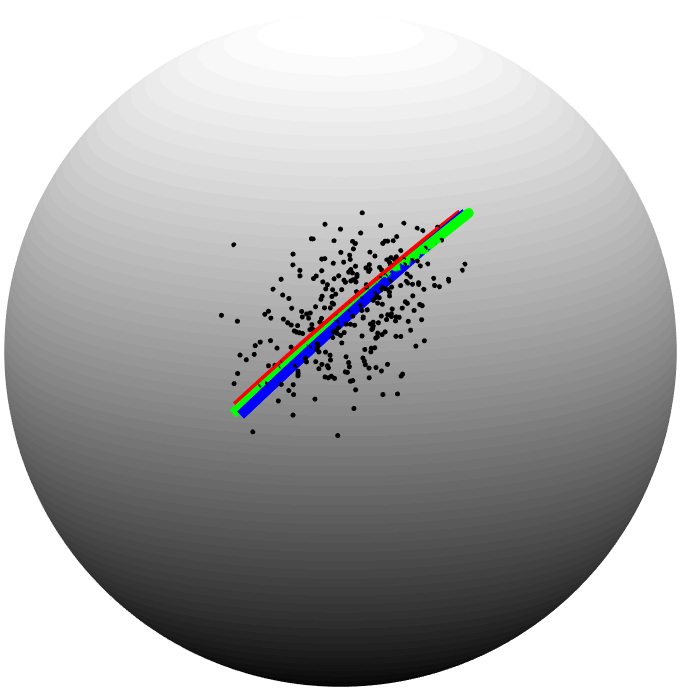}
}
\hspace{2cm}
\centering
\subfigure[]{\label{fig:sphere_linear}
\includegraphics[width=0.3\columnwidth]{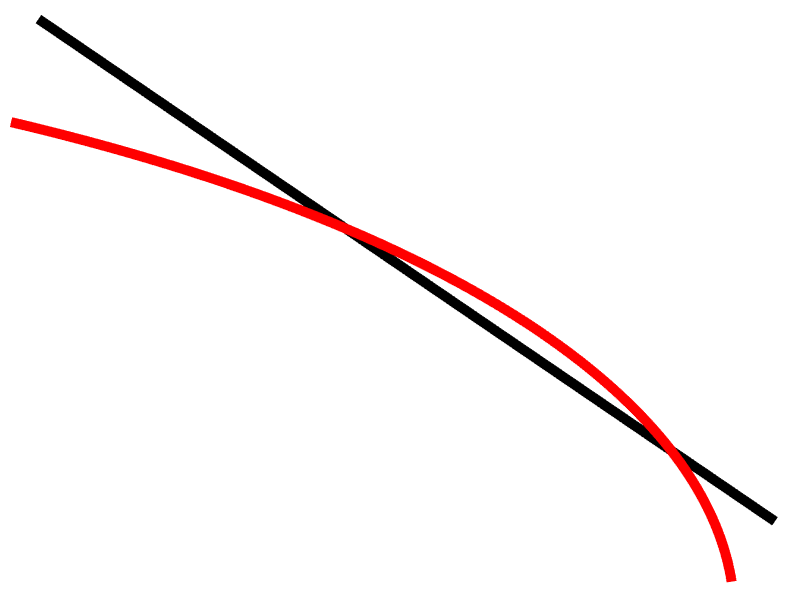}
}
\caption{(a): Geodesic regression on sphere using BGRM and PPGA model. The blue line is the ground truth geodesic. The red line is the estimated geodesic of BGRM model. The green line is the estimated geodesic of PPGA model. (b): The comparison of estimated geodesic of BGRM and PCA model. The red line is the estimated geodesic of BGRM in (a). The black line is the estimated geodesic using PCA, which is a straight line (below the sphere) and cannot be visualized in (a).  }
\label{fig:sphere_re}
\end{figure}

In Table \ref{tab:my_label}, the recovered parameters $\mu, v$, and $\tau$ of our BGRM model are closer to ground truth than three baseline methods (PPGA, PGA and PCA). We can visualize the estimated geodesic in Fig.~\ref{fig:sphere}.  The blue line is the true geodesic. The red line is the estimated geodesic of BGRM model, and the green line is the estimated geodesic of PPGA model. Although both the PPGA and BGRM model can recover the geodesic, our BGRM model kills the unnecessary dimensionality of $v$. The true $v$ has the size of $3 \times 2$, and our BGRM model kills the second column, which is much smaller than the first column. This result demonstrates the ability of our model in automatical dimensionality selection.

To show the correctness of our model, we also compare the estimation results of our model with PCA in the Euclidean space. As shown in Fig. \ref{fig:sphere_linear}, the estimated geodesic of BGRM model is a curve on the sphere, but the estimated geodesic of PCA is a straight line, which is below the sphere. And this leads to the unobserved PCA results on the sphere.

\begin{figure}[t]
\centering
\subfigure[Shape variations of pentagon]{\label{fig:five}
\includegraphics[width=0.5\columnwidth]{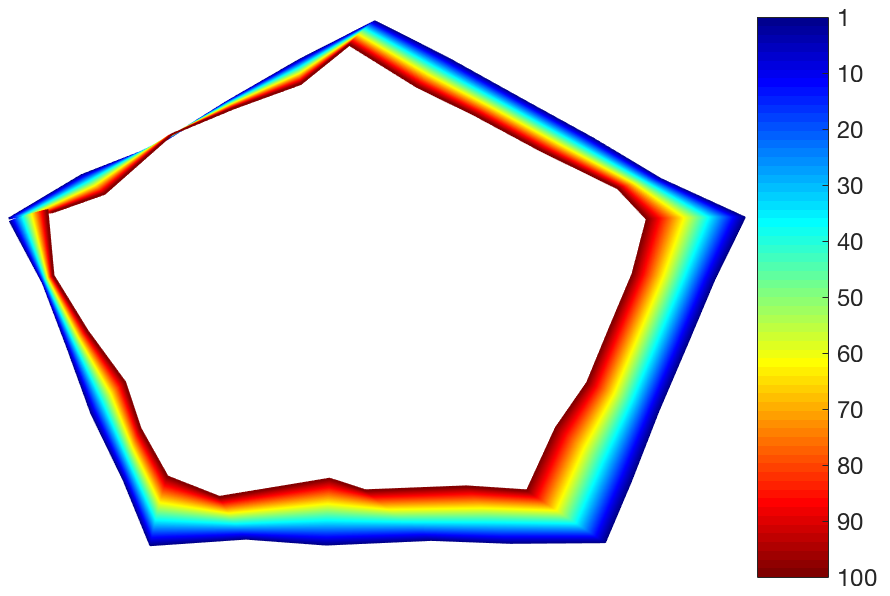}
}
\centering
\subfigure[Reduced dimensionality]{\label{fig:five_dim}
\includegraphics[width=0.45\columnwidth]{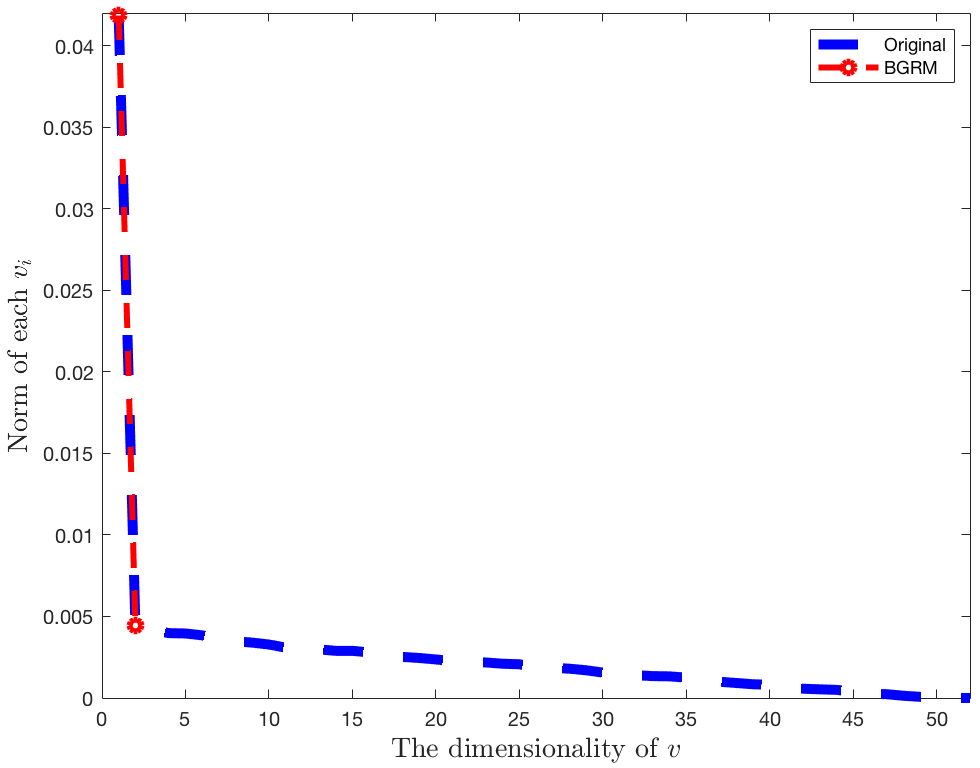}
}
\caption{The shape variations of pentagon and reduced dimensionality using BGRM model. (a): The color from blue to red is the corresponding $x$ of each shape. The shape shrinks with the increasing of $x$. (b): BGRM model automatically selects the first two dimensionalities out of 52 in total.  }
\label{fig:five_all}
\end{figure}

{\bf Pentagon.} To evaluate our model in a high dimensional data, we first apply the BGRM model in analyzing shape variations of synthetic pentagon dataset. This synthetic data contains a collection of 50 pentagon shapes with $x$ from 1 to 50. Each pentagon has the $2 \times 26 $ points. We aim to predict the new pentagon shapes when $x$ is from 51 to 100. The regression result is shown in Fig. \ref{fig:five}. The pentagon shape shrinks with the increasing of the $x$. This example illustrates our model is applicable in analyzing the shape variations of data. In addition, Fig.~\ref{fig:five_dim} shows that our BGRM model reduces the dimensionality of pentagon data from $52$ to $2$. The red dash line is the reduced number of the tangent vector ($v$), and the blue dash line is the original number of $v$.

\begin{figure}[H]
\centering
\includegraphics[scale=0.3]{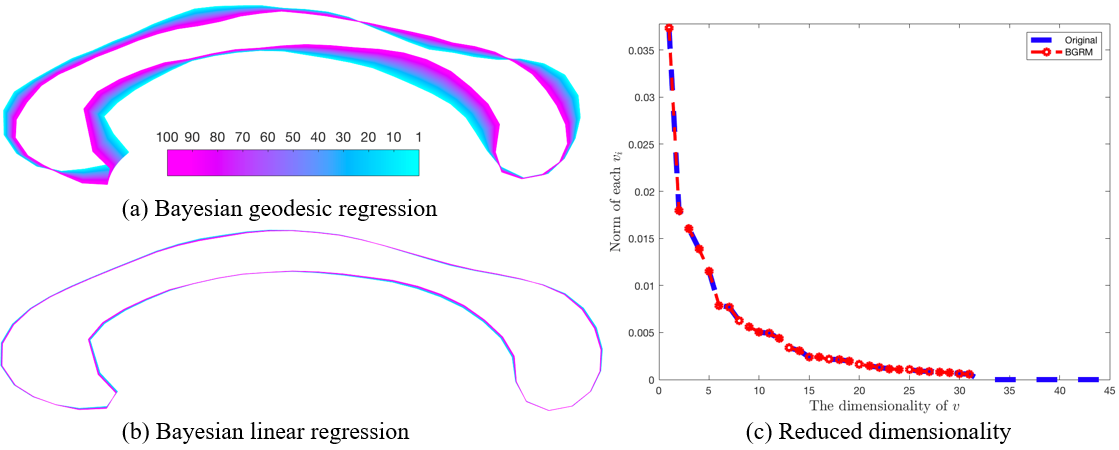}
\vspace{+0.3cm}
\caption{The regression results comparison of BGRM (a) and BLR (b) model using corpus callosum data. The estimated shapes are shown as the sequence from 1 (cyan) to 100 (pink). The color bar indicates the age in years. (c): The reduced dimensionality of BGRM model. }
\label{fig:cc}
\end{figure}

{\bf Corpus Callosum Aging.} To show the effectiveness of BGRM model in the real shape changes, we use corpus callosum data, which are extracted from the MRI scans of human brains.  This data contains a collection of 40 shapes with age from 0 to 80 years. Each corpus callosum shape has the $2 \times 65 $ points. As shown in Fig.~\ref{fig:cc}, we show the corpus callosum shape changes using the BGRM and BLR model. From Fig. 5(a), there are more changes with the increasing of age in the posterior part of the corpus callosum. This result is significantly 
better than the regression results in BLR model, in which shapes are almost overlaying with each other due to the shape geometry does not consider in the Euclidean space of BLR model. Also, as shown in Fig. 5(b), our BGRM model reduces the dimensionality of corpus callosum data from $130$ to $31$. Dimensionality is omitted in Fig. 5(c) if it is greater than 45.


\begin{figure}[h]
\centering
\subfigure[Estimated mandible shapes]{\label{fig:mandible_re}
\includegraphics[width=0.4\columnwidth]{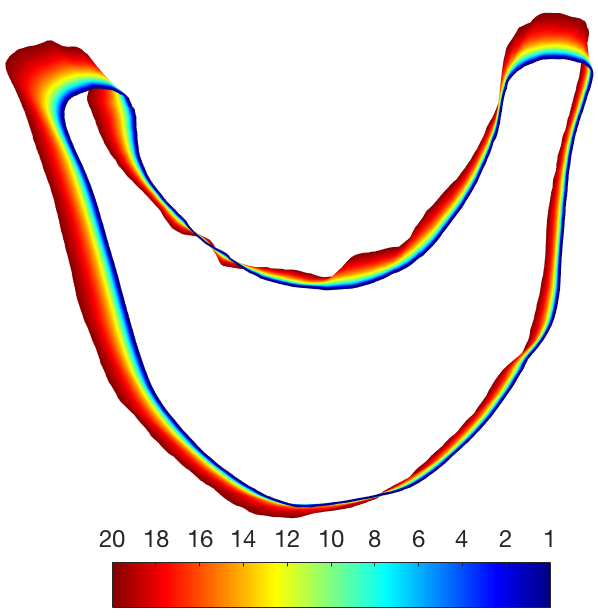}
}
\centering
\subfigure[Reduced dimensionality comparison]{\label{fig:mandible_dim}
\includegraphics[width=0.5\columnwidth]{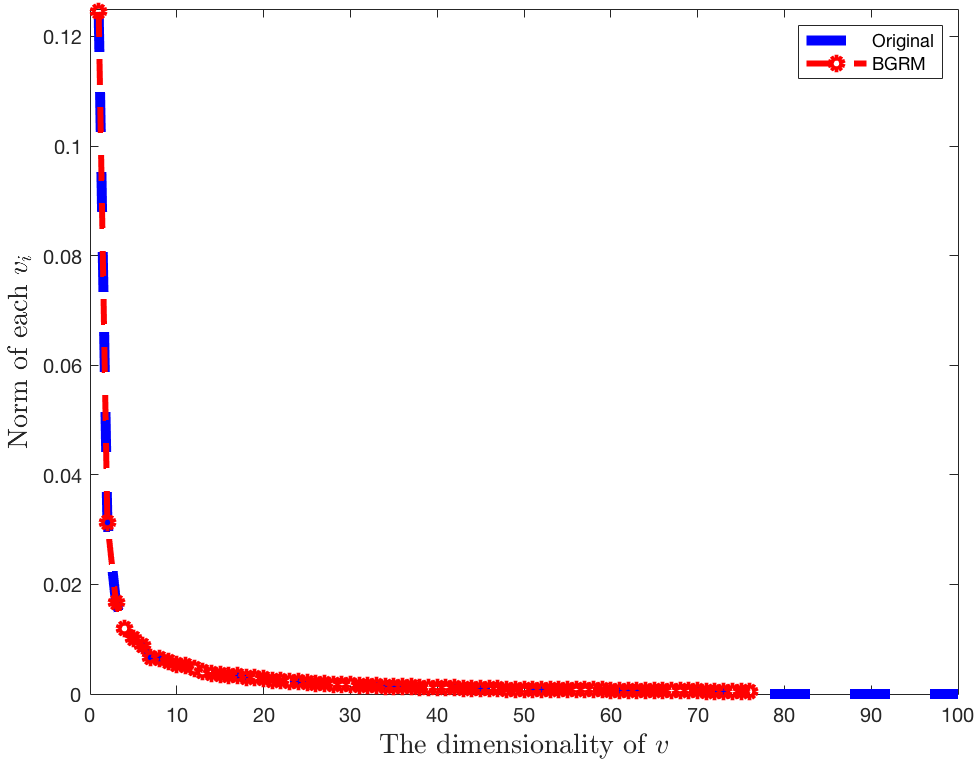}
}
\caption{(a): The estimated human mandible shape using BGRM; (b): The comparison of original dimensionality and reduced dimensionality of BGRM model. }
\label{fig:mandible_results}
\end{figure}

{\bf Mandible shape.} We also evaluate our BGRM model in estimating the shape variations of human mandible growths. The mandible data are extracted from a collection of CT scans of human mandibles. It contains $77$ subjects and the age is from $0$ to $19$ years. We sample $2 \times 400$ points on the boundaries. Fig.~\ref{fig:mandible_re} shows the estimated mandible shape variations using BGRM model, there are more variations in the head part of mandible shape. When the age is between 14 and 20 years, the mandible is dramatically increased, which coincides with the results in~\cite{sharma2014age}. As shown in Fig.~\ref{fig:mandible_dim},  our BGRM model reduces the dimensionality of mandible data from $800$ to $76$. Dimensionality is omitted in Fig.~\ref{fig:mandible_dim} if it is greater than 100. 


\begin{table}[h]
\small
\begin{center} 
\caption{$R^2$ statistic of predicting shapes}
\vspace{+0.3cm}
 \setlength{\tabcolsep}{+2.3mm}{
\begin{tabular}{ccccccccc|c|c|c|}
\hline \label{tab:R2}
Datasets &  Pentagon & Corpus callosum  & Mandible \\
\hline 
Linear regression & 0.0135  & 0.0194  & 0.0518\\
Geodesic regression~\cite{fletcher2011geodesic}  & 0.0223  & 0.0234  & 0.0873\\
ShapeNet~\cite{zhang2019shapenet}   & 0.3911  & 0.3854  & 0.1738\\
\hline
\hline
\textbf{BGRM}  & \textbf{0.4318}  & \textbf{0.4279}  & \textbf{0.2146}\\
\hline
\end{tabular}}
\end{center}
\vspace{-.2in}
\end{table}

\subsection{Significance analysis}

To show the significance of our model, we report the $R^2$ value in Eq.~\eqref{eq:r2}, which  is between $[0, 1]$. The higher the $R^2$ value is, the more variations are explained by the model, and the better of model in predicting the shape.
\begin{equation}\label{eq:r2}
    R^2=1-\frac{\text{Unexplained variation }}{\text{Total variation}}=1-\frac{\sum_{i=1}^N (y_{i} - y_{i}')^2}{\sum_{i=1}^N  (y_{i} - \Bar{y})^2},
\end{equation}
where $y_{i}'$ is the predicted shape, $\Bar{y}$ is the mean shape of $Y$ and $N$ is the number of shapes.

Table~\ref{tab:R2} compares the $R^2$ statistic of our BGRM model with linear regression, geodesic regression model and ShapeNet. The $R^2$ values of three datasets from our BGRM model are larger than that of other models. The lower value of the geodesic regression model means that shape variability is not well modeled by age since age only describes a small fraction of the shape variation. Other factors (gender, weight, etc.) can also affect shape changes. However, the coefficient of determination ($R^2$ values) of our model demonstrates that age is an important factor that affects these shapes. Therefore, our BGRM model is better than state-of-the-art methods and effective in predicting shape variations.

\section{Discussion}
From the above experiments, we find that the proposed BGRM model is able to predict the shape changes with a higher $R^{2}$ value. Although the $R^2$ value of the mandible dataset is not as significant as the other two datasets, which is caused by little variations of the original data, BGRM model still has a higher $R^2$ value than three baseline methods. In addition, we also calculate the p-values of three datasets (0.7485, 0.9780, and 0.3478, respectively). These results imply that predicting shapes are similar to true shapes. However, one weakness of our model is that it is sensitive to large changes in shape.

\section{Conclusion}
In this paper, we develop a Bayesian geodesic regression model on Riemannian manifolds. By introducing a prior to the geodesic regression model, we can automatically select the number of relevant dimensions by driving unnecessary tangent vectors to zero.  We use maximum a posterior method to estimate model parameters.  Four experimental results indicate that our BGRM model takes the advantages of automatically reducing the dimensionality of the subspace and shows the reasonable shape variations. There are some obvious next steps. More data can be applied in our model. The model can also be extended to develop a Bayesian poly-nomial geodesic regression model on Riemannian manifolds.

\small
\bibliography{egbib}
\end{document}